\documentclass[conference]{IEEEtran}
\IEEEoverridecommandlockouts
\usepackage{cite}
\usepackage{amsmath,amssymb,amsfonts}
\usepackage{algorithmic}
\usepackage{graphicx}
\usepackage{textcomp}
\usepackage{xcolor}
\usepackage{subfig}
\usepackage{float}
\def\BibTeX{{\rm B\kern-.05em{\sc i\kern-.025em b}\kern-.08em
    T\kern-.1667em\lower.7ex\hbox{E}\kern-.125emX}}
\begin{document}

\title{Low cost enhanced security face recognition\\ with stereo cameras
}

\author{\IEEEauthorblockN{Biel Tura Vecino}
\IEEEauthorblockA{\textit{Universitat Politècnica de Catalunya}\\
\tt biel.tura@upc.edu}
\and
\IEEEauthorblockN{Martí Cobos}
\IEEEauthorblockA{\textit{Idneo Inc.} \\
\tt marti.cobos@idneo.com}
\and
\IEEEauthorblockN{Philippe Salembier}
\IEEEauthorblockA{\textit{Universitat Politècnica de Catalunya} \\
\tt\small philippe.salembier@upc.edu}
}

\maketitle

\begin{abstract}
This article explores a face recognition alternative which seeks to contribute to resolve current security vulnerabilities in most recognition architectures. Current low cost facial authentication software in the market can be fooled by a printed picture of a face due to the lack of depth information. The presented software creates a depth map of the face with the help of a stereo setup, offering a higher level of security than traditional recognition programs. Analysis of the person's identity and facial depth map are processed through deep convolutional neural networks, providing a secure low cost real-time face authentication method.
\end{abstract}

\begin{IEEEkeywords}
Computer Vision, Stereo cameras, 3D reconstruction, Facial recognition
\end{IEEEkeywords}

\section{Introduction}
Actual face recognition algorithms based on deep neural networks trained on millions of images are believed to be rapidly approaching human-level performance. Face recognition has become the current trend for bio-metric authentication, surpassing the most typical fingerprint verification. Yet, facial recognition, as an authentication method, remains relatively insecure due to the fact that most systems use a unique sensor and are not able to distinguish between a person and a photograph of the subject. This paper presents a secure and low cost alternative to current facial recognition systems based on the extraction of depth information.  

Nowadays, there are mainly two techniques used for extracting depth information: stereo cameras and infrared (IR) dot projector. Dot projector with infrared light involves using a structured light source to capture 3D images. Structured light sources project a known pattern of light onto the surfaces of a physical object to detect deformations in it. Using the captured pattern deformities through a depth sensor, the system is able to compute the surface information. Sensors required for this technique are not cheap. The processing of the sensor's outputted data is not trivial and it requires proprietary libraries to deal with the 3D data. Stereo systems generates a depth map of the scene by capturing two images. This allows the camera to simulate human binocular vision, and therefore gives it the ability to capture three-dimensional properties. The depth information is extracted by matching the distance between the same space point within the two captured images. In many other cases, face recognition doesn't require the most advanced camera and thus, two cameras is, in most cases, cheaper than an infrared dot projector setup.

\section{Theoretical stereo description}
This section presents a generic mathematical notation used to formulate the computation of the depth map though a stereo setup which it's widely used throughout the literature \cite{stereo_depth_map}. This description of the depth map computation assumes that both images from different points of view are aligned and, therefore, both intrinsic and extrinsic camera parameters have been correctly calibrated \cite{distortion, plane_calibration}.

A captured image is defined as $I = (x,y)$ where each $(x,y)$ represents the pixel position within the image. Let us consider two images $I_1, I_2$ captured from two different calibrated cameras. Images aligned in the same plane can be used to compute the disparity between pixel lines, giving a relation of the depth information in the scene. Those lines of disparity matching are called epipolar lines. As Szeliski described in his book \cite{Szeliski}, the epipolar geometry for a pair of cameras is implicit in the relative pose of the stereo setup.

A comparison criteria must be used in order to determine the disparity between the two images. The main assumption in stereo corresponding is that a pixel $(x,y)$ on one image $I_1$ must have its correspondent pixel $(x', y')$ in the other image $I_2$ such that $(x',y') = (x - d_x, y - d_y)$ where $d_x$ and $d_y$ are the disparity values on each direction. The depth values $z$ in the depth map are inversely proportional to the absolute disparity $d = |d_x + d_y|$. Through some trigonometrical relations with the physical distance between two cameras $T$ and the focal length $f$, the relation between the disparity in both images and the real distance of that point follows:
\begin{equation}
    \frac{T-d}{z-f} = \frac{T}{z} \implies z = \frac{f T}{d}
\end{equation}

\begin{figure}[ht]
\begin{center}
   \includegraphics[width=\linewidth]{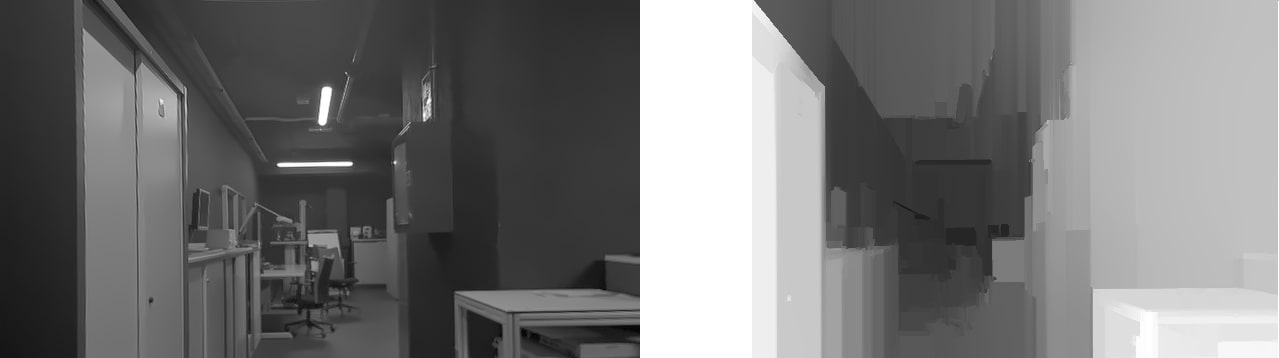}
\end{center}
   \caption{\small Depth map generated through the stereo camera setup following the theoretical stereo description model and mapping $z$ values to a gray-scale image.}
\label{fig:depth_map}
\end{figure}

Mapping the disparity to a gray-scale image generates a map where depth is proportional to brightness (Figure \ref{fig:depth_map}).

\subsection{Face modeling}
\begin{figure}[ht] 
    \begin{center}
    \subfloat[Original depth map]{%
        \includegraphics[width=0.18\textwidth]{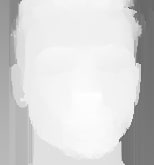}%
        \label{fig:original_depth}%
        }%
    \hspace{0.05\textwidth}
    \subfloat[Modeled depth map]{%
        \includegraphics[width=0.18\textwidth]{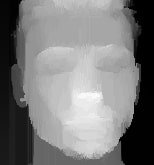}%
        \label{fig:gamma_depth}%
        }%
    \caption{\small Depth map generated for a real person's face}
    \end{center}
\end{figure}

The mentioned stereo procedure estimates the depth values of all the image captured by the camera. However, it is clear that the depth distance of the subject face features is, in most cases, smaller than the depth distance between the subject and the background. This results into a depth map where face deepness cannot be easily distinguished, resulting in a face depth map that looks like an homogeneous white area (Figure \ref{fig:original_depth}). Being not able to differentiate the depth information does not mean it is not present in the captured depth map.

Facial depth information can be improved by enhancing the contrast. Through a power law function, which expands or compress bright pixels, we can retrieve this computed depth information. The resulting image is shown in in Figure \ref{fig:gamma_depth}. After the applied pixel transformation, the image changes from a very low-detailed depth map to a map in which face profundity can be easily distinguished.

\section{Proposed method}

\begin{figure*}[t]
\begin{center}
\includegraphics[width=0.95\linewidth]{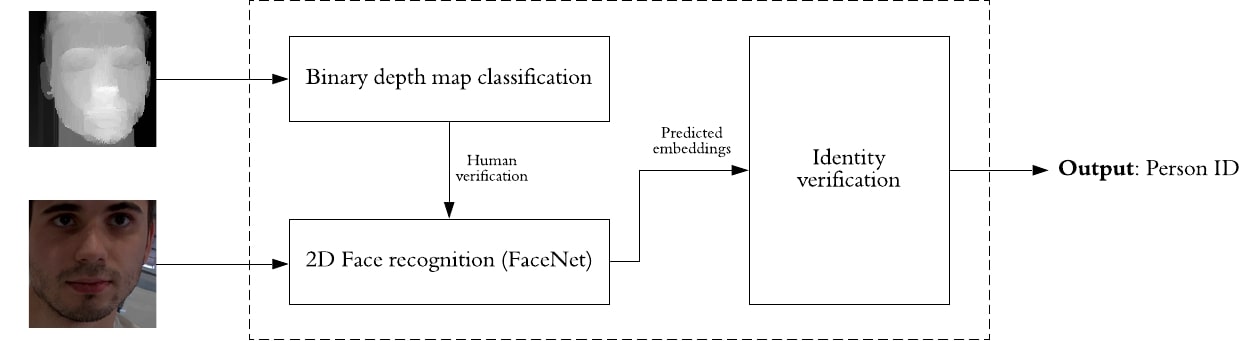}%
\end{center}
   \caption{\small Block diagram of the proposed facial recognition software with the addition of the depth map classifier. The input of the recognition software is the RGB image of the person's face and the generated gray-scale depth map image.}
\label{fig:scheme}
\end{figure*}

Our software is based on the well known FaceNet architecture \cite{faceNet} trained on the 2D image dataset of Labelled Faces in the Wild (LFW) \cite{LFWTech}. The added value of our proposed method is the possibility to feed the depth map and add a level of security to this architecture. A block diagram of the proposed scheme is shown in Figure \ref{fig:scheme} below. The input of the proposed recognition system consists on two different images: a gray-scale face depth map and a RGB face. The depth map is parsed through a binary classification to check whether the depth map corresponds to a real human or not. The RGB face is the input of the 2D Face Recognition and the output of the network is, as most of current facial recognition deep learning models, a vector that represents the face embeddings.

\subsection{Binary depth map classification}
Although three-dimensional facial recognition is gaining popularity, literature survey shows that there is no deep convolutional neural network designed specifically for this purpose. The main reason for this is the lack of huge amounts of 3D training and test data \cite{3d_dataset}. To enhance the security of current facial recognition models available, an alternative solution is proposed in this article: a simpler convolutional neural network is designed to solve a binary classification problem. This type of network is used to classify the input data in positive or negative samples. In this case, a positive image will reefer to depth map of a real face and a negative image otherwise.

The model designed for this task is a convolutional neural network with one single output in the last layer. The input of the network is a $96 \times 96 \times 1$ gray-scale image of the depth map. The output gives the confidence of the input image being a real face depth map. The proposed model consists of 3 convolutional layers with 3 pooling layers and 3 fully-connected layers. The network's architecture is described in Table \ref{table:cnn}.

\begin{table}[h]
    \begin{center}
        \begin{tabular}{|l|c|c|c|}
        \hline
        \textbf{Layer} & \textbf{Size-in} & \textbf{Size-out} & \textbf{Kernel} \\
        \hline\hline
        conv1 & $96 \times 96 \times 1$ & $96 \times 96 \times 8$ & $5 \times 5$ \\
        \hline
        pool1 & $96 \times 96 \times 8$ & $32 \times 32 \times 8$ & $3 \times 3$ \\
        \hline
        conv2 & $32 \times 32 \times 8$ & $32 \times 32 \times 16$ & $3 \times 3$ \\
        \hline
        pool2 & $32 \times 32 \times 16$ & $16 \times 16 \times 16$ & $2 \times 2$ \\
        \hline
        conv3 & $16 \times 16 \times 16$ & $16 \times 16 \times 32$ & $3 \times 3$ \\
        \hline
        pool3 & $16 \times 16 \times 32$ & $8 \times 8 \times 32$ & $2 \times 2$ \\
        \hline
        fc1 & $8 \times 8 \times 32$ & $1 \times 128$ & \\
        \hline
        fc2 & $1 \times 128$ & $1 \times 32$ & \\
        \hline
        fc3 & $1 \times 32$ & $1 \times 1$ & \\
        \hline
        \end{tabular}
    \end{center}
    \caption{\small Binary classification CNN structure}
    \label{table:cnn}
\end{table}
        
The last output is activated through a Sigmoid function to map the output value to a probability that expresses the confidence of the network about it being a real person depth map. The other layers use ReLu as their activation functions.

\subsection{Dataset and training}
The network is trained using a Binary Cross-Entropy Loss function, \textit{BCE}. This loss function measures the performance of a classification model whose output is a single probability value between 0 and 1. The optimizer used to reduce the loss of the network is Adam \cite{adam}.

\begin{figure}[!ht]
\begin{center}
   \includegraphics[width=0.85\linewidth]{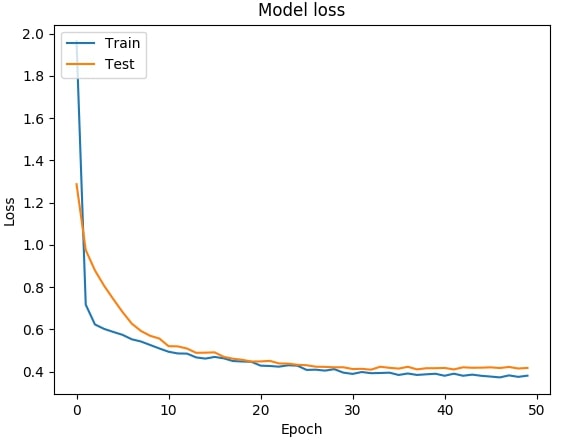}
\end{center}
   \caption{\small Loss evolution through the training of the binary depth map classification network.}
\label{fig:loss}
\end{figure}

The used dataset for training consists on approximately 2200 scans of 35 different persons and 1200 scans of 10 different printed identities and 20 different scenes. The network achieves 86.8\% accuracy in the training dataset and 82.2\% in the validation dataset (Figure \ref{fig:loss}).

\subsection{Threshold definition}
A threshold is defined to set the minimum confidence of the depth map network to decide whether its input is a real face or not. It is important to choose a threshold where most of the predicted classification errors are real faces being recognized as fake than non-faces depth maps being recognized as real.

\begin{figure}[ht]
\begin{center}
   \includegraphics[width=0.90\linewidth]{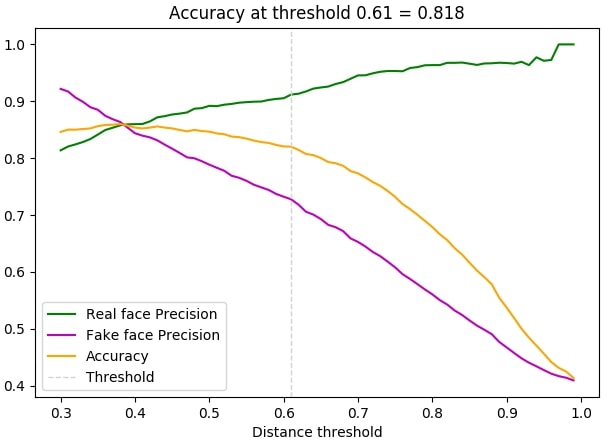}
\end{center}
   \caption{\small Accuracy and Precision of the depth map network for different threshold values $\in (0,1)$}
\label{fig:acc}
\end{figure}

The threshold chosen is $0.61$. The accuracy in our generated dataset at this threshold ($81.8\%$) is shown in Figure \ref{fig:acc} and is lower than the original accuracy with a threshold at $0.5$. From our point of view, it is not a big issue to check the face again if the network is not confident enough. Contrarily, it is very important that the proposed architecture never predicts a real face when a non-face depth map is being recognized. Although the accuracy of the depth map classification at this threshold is not very high for a binary classification task, most mistakes are real faces being recognized as fake and not the other way. As we shall see, a printed photograph will not likely enter the system.

\section{Results}
\begin{figure}[ht] 
    \begin{center}
    \subfloat[Captured image of face and its depth map]{%
        \includegraphics[width=0.35\textwidth]{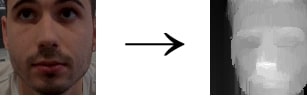}%
        \label{fig:real_depth}%
        }%
    \hspace{0.01\textwidth}
    \subfloat[Captured image of a photography and its depth map]{%
        \includegraphics[width=0.35\textwidth]{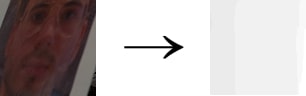}%
        \label{fig:paper_depth}%
        }%
    \end{center}
    \caption{\small Depth map generated from the proposed stereo setup for a real person's face and a photography slightly tilted.}
    \label{fig:person_paper_depth}
\end{figure}

The stereo setup described generates a fairly accurate depth map of the person's face. As it can be seen in Figure \ref{fig:person_paper_depth}, a real face depth map does differentiate from a fake one of a printed face. Although it may seem an easy classification task, it gets more complex when the paper is tilted or folded in several directions. The proposed convolutional neural network learns where the depth information is located and the distribution within the face.

Although the developed program runs in real-time, a simple validation dataset was generated in order to get an accurate assessment for the results of this paper. The validation dataset consists of 20 pairs of frontal face images (right and left frames) of 3 different identities in 4 different scenes with a variety of light conditions. This dataset includes pictures of the real person and fake printed faces. 

The results presented evaluate the confidence of the Binary depth map network (D. C.) and the output of the recognition system in four different scenes (Tables \ref{table:face_results_a}, \ref{table:face_results_b}, \ref{table:face_results_c} ans \ref{table:face_results_d}). The output of the software can be the person's \textit{ID} if it has been recognized, \textit{Unknown} if the embeddings do not match any of the extracted feature vectors of the dataset or \textit{None} if the recognition system has not been triggered. Scenes with a uniform illumination and without reflections perform significantly better than the ones with worse lighting conditions. The fact that both cameras capture different light reflection makes obviously the stereo matching less accurate.

\begin{table}[ht]
    \centering
    \begin{tabular}{|c|c|c|c|}
    \hline
    \textbf{ID} & \textbf{Face image type} & \textbf{D.C.} & \textbf{Output} \\ \hline \hline
    1 & {Real registered person} & 0.72 & {\color{red}Unknown} \\ \hline
    2 & {Real registered person} & {\color{red}0.58} & {\color{red}None}\\ \hline
    3 & {Real non-registered person} & 0.67 & Unknown \\ \hline
    1 & {Registered printed face} & 0.43 & None \\ \hline
    2 & {Non-Registered printed face} & 0.36 & None \\ \hline
    \end{tabular}
    \caption{\small Results of the proposed network for a scene with a very textured background with a lot of light reflections. Left camera sees different light sources than right camera.}
    \label{table:face_results_a}
\end{table}

\begin{table}[ht]
    \centering
    \begin{tabular}{|c|c|c|c|}
    \hline
    \textbf{ID} & \textbf{Face image type} & \textbf{D.C.} & \textbf{Output} \\ \hline \hline
    1 & {Real registered person} & 0.75 & ID 1 \\ \hline
    2 & {Real registered person} & 0.79 & ID 2 \\ \hline
    3 & {Real non-registered person} & 0.86 & Unknown \\ \hline
    1 & {Registered printed face} & 0.20 & None \\ \hline
    2 & {Non-Registered printed face} & 0.26 & None \\ \hline
    \end{tabular}
    \caption{\small Results of the proposed system for a well illuminated scene with day light and no reflections. The background of this scene is a low textured white wall.}
    \label{table:face_results_b}
\end{table}

\begin{table}[ht]
    \centering
    \begin{tabular}{|c|c|c|c|}
    \hline
    \textbf{ID} & \textbf{Face image type} & \textbf{D.C.} & \textbf{Output} \\ \hline \hline
    1 & {Real registered person} & 0.89 & ID 1 \\ \hline
    2 & {Real registered person} & 0.81 & ID 2 \\ \hline
    3 & {Real non-registered person} & 0.91 & Unknown \\ \hline
    1 & {Registered printed face} & 0.13 & None \\ \hline
    2 & {Non-Registered printed face} & 0.16 & None \\ \hline
    \end{tabular}
    \caption{\small Results of the proposed system for a room with uniform illumination by a controlled light sources. The background is a simple black wall with minimum texture.}
    \label{table:face_results_c}
\end{table}

\begin{table}[ht]
    \centering
    \begin{tabular}{|c|c|c|c|}
    \hline
    \textbf{ID} & \textbf{Face image type} & \textbf{D.C.} & \textbf{Output} \\ \hline \hline
    1 & {Real registered person} & 0.85 & ID 1 \\ \hline
    2 & {Real registered person} & 0.87 & ID 2 \\ \hline
    3 & {Real non-registered person} & {\color{red}0.55} & {\color{red}None} \\ \hline
    1 & {Registered printed face} & 0.37 & None \\ \hline
    2 & {Non-Registered printed face} & 0.22 & None \\ \hline
    \end{tabular}
    \caption{\small Results of the proposed system for an outdoors scene with direct sun light and a textured background without reflections.}
    \label{table:face_results_d}
\end{table}

To summarize the results presented, a confusion matrix of the proposed system is shown in Table \ref{table:confussion}. The average precision of the system in this validation dataset is $88.75\%$, giving an average F1 score of $84.43 \%$. Note that any fake identity (\textit{None}) did not trigger the recognition system.

\begin{table}[H]
    \renewcommand{\arraystretch}{1.1}
    \centering
    \begin{tabular}{c|c|c|c|c|}
    \cline{2-5}
     & \textbf{ID 1} & \textbf{ID 2} & \textbf{Unknown} & \textbf{None}\\ \hline
    \multicolumn{1}{|c|}{Predicted ID 1} & 3 & 0 & 0 & 0\\ \hline
    \multicolumn{1}{|c|}{Predicted ID 2} & 0 & 3 & 0 & 0\\ \hline
    \multicolumn{1}{|c|}{Predicted Unknown} & 1 & 0 & 3 & 0\\ \hline
    \multicolumn{1}{|c|}{Predicted None} & 0 & 1 & 1 & 8\\ \hline
    \end{tabular}
    \caption{\small Confusion matrix of the face recognition system}
    \label{table:confussion}
\end{table}

\newpage

\section{Discussion}
To sum up, this paper presents a low cost solution for a securer facial recognition system. Through combining classical Computer Vision techniques with the potential and accuracy of deep convolutional neural networks, we have built a system that can be implemented in most of the situations where facial recognition is starting to gain more popularity. The fact that a pair of cameras is cheaper than an infrared dot projector setup and that building a system that understands the 3D points given by the IR projector is more expensive, poses a strong argument for the development of the architecture proposed in this paper.

The generation of the depth map is computationally fast if the resolution of both stereo images is not very large. As the proposed network learns the facial attributes of a stereo depth map presented as a gray-scale images, it is not needed to have a very detailed depth map. In our experiments, we could achieve a recognition pipeline (both stereo matching and recognition) that performed one frame recognition every $0.8$s. In the real case where a person uses the recognition system, we averaged that 2 to 3 frames had to be analyzed. The first frame usually captures the person's in motion, approaching to the system, resulting in blurry RGB picture and an unclear depth map. The second frame is usually correctly recognized and, if not, the third frame always worked in our experiments. This is, at most, a $2.4$s process for a securer facial recognition system.

There are some aspects that could be improved in our model for future revisions of the presented software. The worst results were shown when the background had difficult lighting conditions such as different light reflections. A background modeling of the scene could be performed in order to only generate a depth map of the person's face. By implementing this strategy, the mentioned face modeling would not be necessary and it will immediately eliminate unwanted reflections captured that difficult the correct generation of the depth map.

\bibliographystyle{IEEEtran}
\bibliography{egbib}

\begin{thebibliography}{1}
\providecommand{\url}[1]{#1}
\csname url@samestyle\endcsname
\providecommand{\newblock}{\relax}
\providecommand{\bibinfo}[2]{#2}
\providecommand{\BIBentrySTDinterwordspacing}{\spaceskip=0pt\relax}
\providecommand{\BIBentryALTinterwordstretchfactor}{4}
\providecommand{\BIBentryALTinterwordspacing}{\spaceskip=\fontdimen2\font plus
\BIBentryALTinterwordstretchfactor\fontdimen3\font minus
  \fontdimen4\font\relax}
\providecommand{\BIBforeignlanguage}[2]{{%
\expandafter\ifx\csname l@#1\endcsname\relax
\typeout{** WARNING: IEEEtran.bst: No hyphenation pattern has been}%
\typeout{** loaded for the language `#1'. Using the pattern for}%
\typeout{** the default language instead.}%
\else
\language=\csname l@#1\endcsname
\fi
#2}}
\providecommand{\BIBdecl}{\relax}
\BIBdecl

\bibitem{stereo_depth_map}
O.~Krutikova, A.~Sisojevs, and M.~Kovalovs, ``Creation of a depth map from
  stereo images of faces for 3d model reconstruction,'' \emph{Procedia Computer
  Sc.}, pp. 452--459, 2017.

\bibitem{distortion}
S.~Beauchemin and R.~Bajcsy, ``Modelling and removing radial and tangential
  distortions in spherical lenses,'' \emph{Lecture Notes in Computer Science},
  pp. 1--21, 2000.

\bibitem{plane_calibration}
Z.~Zhang, ``Flexible camera calibration by viewing a plane from unknown
  orientation,'' \emph{Proceedings of the 7th International Conference on
  Computer Vision 99}, pp. 666--673, 1999.

\bibitem{Szeliski}
R.~Szeliski, \emph{Computer Vision: Algorithms and Applications}.\hskip 1em
  plus 0.5em minus 0.4em\relax Berlin, Heidelberg: Springer, 2010.

\bibitem{faceNet}
F.~Schroff, D.~Kalenichenko, and J.~Philbin, ``Facenet: {A} unified embedding
  for face recognition and clustering,'' \emph{CoRR}, 2015.

\bibitem{LFWTech}
B.~Huang, M.~Ramesh, T.~Berg, and E.~Learned-Miller, ``Labeled faces in the
  wild: A database for studying face recognition in unconstrained
  environments,'' in \emph{{Workshop on Faces in 'Real-Life' Images: Detection,
  Alignment, and Recognition}}, 2007.

\bibitem{3d_dataset}
S.~Z. Gilani and A.~Mian, ``Learning from millions of 3d scans for large-scale
  3d face recognition,'' \emph{CoRR}, 2017.

\bibitem{adam}
P.~Kingma and J.~Ba, ``Adam: A method for stochastic optimization,'' \emph{3rd
  International Conference for Learning Representations, San Diego}, 2015.

\end{thebibliography}

\end{document}